\documentclass{article}

\usepackage{arxiv}

\usepackage[utf8]{inputenc} 
\usepackage[T1]{fontenc}    
\usepackage{hyperref}       
\usepackage{url}            
\usepackage{booktabs}       
\usepackage{amsfonts}       
\usepackage{nicefrac}       
\usepackage{microtype}      
\usepackage{lipsum}		
\usepackage{graphicx}
\usepackage{doi}
\usepackage{longtable}

\title{An Appraisal-Based Chain-Of-Emotion Architecture for Affective Language Model Game Agents}

\author{ \href{0000-0002-5125-8277}{\includegraphics[scale=0.06]{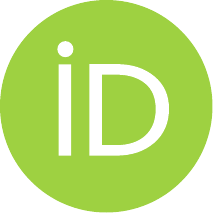}\hspace{1mm}Maximilian Croissant} \\
	Department of Computer Science\\
	University of York\\
	York, UK \\
	\texttt{mc2230@york.ac.uk} \\
	\And
	\href{0000-0003-0807-6395}{\includegraphics[scale=0.06]{orcid.pdf}\hspace{1mm}Madeleine Frister} \\
        Department of Computer Science\\
	University of York\\
	York, UK \\
	\texttt{mf1255@york.ac.uk} \\
	\And
	\href{0000-0003-1115-1018}{\includegraphics[scale=0.06]{orcid.pdf}\hspace{1mm}Guy Schofield} \\
        School of Arts and Creative Technologies\\
	University of York\\
	York, UK \\
	\texttt{guy.schofield@york.ac.uk} \\
 	\And
	\href{0000-0003-0746-8899}{\includegraphics[scale=0.06]{orcid.pdf}\hspace{1mm}Cade McCall} \\
        Department of Psychology\\
	University of York\\
	York, UK \\
	\texttt{cade.mccall@york.ac.uk} \\
}

\date{}

\renewcommand{\headeright}{Preprint}
\renewcommand{\undertitle}{Preprint}
\renewcommand{\shorttitle}{Chain-Of-Emotion for Affective LLM Agents}

\hypersetup{
pdftitle={An Appraisal-Based Chain-Of-Emotion Architecture for Affective Language Model Game Agents},
pdfsubject={CS},
pdfauthor={Maximilian~Croissant, Madeleine~Frister, Guy~Schofield, Cade~McCall},
pdfkeywords={Large Language Models, Affective COmputing, AI Agents, Affect, Simulation, Video Games},
}

\begin{document}
\maketitle

\begin{abstract}
The development of believable, natural, and interactive digital artificial agents is a field of growing interest. Theoretical uncertainties and technical barriers present considerable challenges to the field, particularly with regards to developing agents that effectively simulate human emotions. Large language models (LLMs) might address these issues by tapping common patterns in situational appraisal.  In three empirical experiments, this study tests the capabilities of LLMs to solve emotional intelligence tasks and to simulate emotions. It presents and evaluates a new chain-of-emotion architecture for emotion simulation within video games, based on psychological appraisal research. Results show that it outperforms standard LLM architectures on a range of user experience and content analysis metrics. This study therefore provides early evidence of how to construct and test affective  agents based on cognitive processes represented in language models.
\end{abstract}


\section{Introduction}
In user-centered software and video games, affective artificial agents have long been researched for their potential to provide personalized, natural, and engaging experiences for both entertainment and training purposes \cite{picard2000affective, hudlicka2011guidelines, reisenzein2013computational}. Affective artificial agents are mainly defined through their ability to simulate appropriate emotional responses given certain situations \cite{beale2009affective, hudlicka2011guidelines}. Consequently, they are believed to contribute to enriched user interactions on various domains \cite{beale2009affective}, and even potential health benefits \cite{luxton2022intelligent}.

However, building systems that successfully model, and express emotions is a difficult task \cite{picard2000affective}, especially since affect and emotion are fuzzy concepts, even in psychology research \cite{scherer2022theory}. Affective processes are very complex and involve multiple components (such as cognitive, behavioural, physiological, and feeling \cite{scherer2000psychological}) that are not fully understood and often debated on a fundamental level \cite{izard2010TheRegulation}, which makes computational representations very difficult \cite{hudlicka2011guidelines}. With the development of modern technology, such machine learning \cite{siam2022deploying}, emotion simulation might be achievable through data-driven techniques.  

For example, large language models (LLMs) have demonstrated a potential to simulate a range of cognitive abilities \cite{binz2023using} and even imputing a range of mental and affective states to others \cite{kosinski2023theory}. Because LLMs are trained on large text bodies that hold representations of human mental abilities, they have been observed to exhibit human-like performance on a variety of tasks \cite{binz2023using, hamalainen2023evaluating}. Since emotions are an important part of how humans perceive reality and therefore construct language \cite{barrett2017emotions} and are heavily influenced by cognitive processes \cite{moors2013appraisal} including linguistic labelling \cite{lindquist2015does, brooks2017role}, language-based emotion representations might too enable deep learning models to better simulate affective responses. As of yet, the potential for LLMs to solve some of the issues present in the field of affective agents is however not well understood.

This study therefore examines the potential of LLMs to simulate emotions and how this potential might be influenced by the underlying implementation architecture. Using contemporary findings of emotion research, we propose a cognitive appraisal-based approach for language model affect generation and test it against other strategies in the ability to generate appropriate situational emotions. We then use those results to implement affective agents within a newly developed conversational video game. This study therefore represents an effort to progress affect simulation for artificial agents using language models.

\section{Related Work}

\subsection{Affective Agents}

In affective computing, researchers and developers are interested in creating affective systems that intelligently respond to changes in users' emotions \cite{picard2000affective}. Some of the benefits associated with affective computing techniques applied to video games include more consistent, accessible player experience for a range of different players \cite{gilleade2005affective}, personalized health and training applications \cite{anolli2010emotions}, as well as new and purposefully designed gameplay mechanisms aimed at reinforcing target experiences \cite{hudlicka2008affective, yannakakis2014emotion}. The use of affective agents in video games has been researched with special regard to this last aim. In 2011, Hudlicka discussed potential system design elements for affective (or more precisely emotional) game agents \cite{hudlicka2011guidelines}. According to the author, affective agents can be seen as computational representations of operationalizations made from emotion-theoretical models with appraisal functionality for emotion generation. For example, artificial agents could implement computational calculations of certain events to assess the relevance to the agent and consequently probable emotional reaction \cite{hudlicka2011guidelines}. "Computation-friendly" appraisal implementations have often been built on models such as the OCC model \cite{ortony1990cognitive} (see for example GAMYGDALA \cite{popescu2013gamygdala}). Taken specific fixed aspects (such as expectations of the agent \cite{bosse2009there}) into account, such models have been used to simulate appraisal based on decision trees.

The main aim for such artificial agents is seen to be natural, human-like behaviour and believability as a part of a more fleshed-out and engaging game world \cite{hudlicka2008affective, hudlicka2011guidelines}. The tasks of agents therefore differ from other affective game mechanism that mostly try to adapt the game world to player affect \cite{yannakakis2011experience, croissant2023systematic}. Procedurally generated content (PCG) based on affective information has been shown in video games to successfully increase enjoyment and offer personalized, immersive experience (see for example the work of Shaker et al. \cite{shaker2010towards, shaker2013fusing}). This is often done by fine-tuning certain mechanics shown to be associated with a target player emotion to increase the probability for that emotion \cite{yannakakis2011experience}. Affective agents however do not need to adapt behaviours to player emotions, but rather need their own emotion representations that could then lead to believable behaviours (or other natural representations of emotion components, such as simulated feeling or simulated physiology \cite{hudlicka2011guidelines}). 

The central issues of designing and developing affective game agents lies therefore in creating good computational simulations of emotional states. Human emotions are complex psycho-physiological states that are expressed within behavioural, physiological, cognitive, and feeling components \cite{scherer2022theory}.Moreover, while much work has been done to empirically investigate emotions, many core theoretical disagreements remain \cite{scherer2022theory}, including debates between dimensional \cite{russell1980circumplex}, discrete \cite{izard1992basic}, constructivist \cite{barrett2017emotions}, and cognitive \cite{scherer2000psychological} perspectives. 

A fully developed affective agent would make it necessary to first solve all fundamental psychological gaps that had been present since the beginning of affective computing \cite{picard2000affective} and then integrate them into working, computational systems \cite{hudlicka2011guidelines}. This means that building a psychology-based, fully functional and accurate emotion simulation for an artificial agent is currently not possible and would be in almost all game design cases impractical. 
However, we may still be able to build affective agents that possess key features of emotion elicitation in humans and, as a consequence, allow for relatively successful simulation of human emotions. Once candidate feature is appraisal. Emotion elicitation is dependent on contextual and individual factors, processed through appraisal \cite{moors2013appraisal, scherer2001appraisal}. The notion of emotion appraisal is that emotions are caused by subjective evaluations of triggering events in regards to their significance to one’s personal life or interests \cite{lazarus1991emotion}. Evidence suggests that appraisal holds a central role in emotion elicitation and as a consequence acts on all other emotion components \cite{scherer2001appraisal}. 

Any given external (e.g. situations) or internal (e.g. thoughts) event may be appraised on multiple variables that contribute to emotion forming. Such variables might include goal relevance, certainty, coping potential, or agency \cite{moors2013appraisal}. Appraisal therefore represents a flexible process that adapts to individual differences \cite{gross2003individual} and the current context \cite{mesquita2014emotions}. Evidence also suggests that language can play a key role in emotional appraisal, both by providing key contextual information from which to construct the appraisal and by providing conceptual labels for the appraised states \cite{brooks2017role}.  With all this in mind, language models might provide one mean of simulating the appraisal process as they are able to generate high-level meaning-driven text outputs based on text training data that potentially holds implicit representations of human psychological processes \cite{binz2023using}.  

\subsection{Language Model Approach}

In the last few years, Natural Language Processing (NLP) has been rapidly progressing to the point that singe task-agnostic language models perform well in a range of tasks \cite{brown2020language}, including the simulation of human-like behaviour \cite{hamalainen2023evaluating}. The basis for this is the large amount of training data representing a wide range of human behaviour through language \cite{brown2020language, bommasani2021opportunities}. In relation to games, models such as OpenAI's Generative Pre-trained Transformer (GPT-2 and its successors) have shown early successes in the procedural generation of interactive stories \cite{freiknecht2020procedural}, text-based adventure dialog as well as action candidates \cite{yao2020keep, callison2022dungeons}. In a recent simulation study by Park et al. \cite{park2023generative}, language models were implemented in artificial agent architectures to populate a sandbox world reminiscent of The Sims \cite{sims}. The architecture includes storing and retrieving information from a memory system and based on relevancy for the current situation and then uses the information to generate reflections (i.e. high-level interpretations of a situation), plans (i.e. potential future actions), and immediate actions. Multiple agents were simulated in a game-like world and the authors suggest that emerging interactions were natural and believable in terms of human-like simulation.

While work in this area is still in an early stage, the use of language models addresses some concerns with prior approaches. Most notably, instead of trying to build computational representations of human behaviour, the main task involves trying to retrieve believable human behaviour given a situation from a language model and implementing the results within a game agent. Depending on the game aim, this involves (1) translating the current situation with regards to the expected outcome into language; (2) generating content using a large language model; and (3) translating the output back in order to implement it in a game system. For example, Ciolino et al. \cite{ciolino2020go} used a fine-tuned GPT-2 model to generate Go moves by translating the current board state to text and the language output back to action suggestions. Such a process is naturally easier for purely text-based tasks, such as dialog generation, where text is already the expected output and the expected output can comparatively easily be described in language \cite{yao2020keep, park2023generative}.

Still, even purely text-based generative tasks can pose some potential barriers for language models. The most obvious barrier comes from the underlying training data. No language model can represent human behaviour in its entirety, but is limited to the training data and its biases \cite{sun2022bertscore} as well as model specifications \cite{brown2020language}. Additionally, performance of language model is not only dependent on the output, but also on the input \cite{von2023transformers}. For example, chain-of-thought prompting is a concept introduced by Wei et al. from the Google Research team \cite{wei2022chain} and relates to the integration of chain-of-thought in few-shot prompts to improve reasoning capabilities of language models. Similarly, as Park et al. \cite{park2023generative} describe, simulating believable behaviour in a digital world includes various important steps (including storing and retrieving memory, reflecting in addition to observing, etc.) that ultimately work together to improve the probability to generate expected and natural behaviour. 

When it comes to designing affective agents (i.e. agents that simulate emotions), the first questions that we have to ask is how well is affect represented in the training data and is a language model capable of retrieving it? In a recent paper discussing performance of different GPT iterations on theory of mind tasks, Kosinski \cite{kosinski2023theory} found that new language models perform very well when it comes to imputing unobservable mental and affective states to others. Such findings (especially combined with findings suggesting good performance on cognitive tasks \cite{binz2023using, singh2023mind}) suggest that high-level psychological mechanisms are represented in language alone and could therefore be simulated with a well-constructed language model. Along these linses, can we effectively and efficiently achieve accurate and natural affect-simulation using language models? If we can assume that emotions are represented in language models, mechanisms for emotion elicitation (such as appraisal) might also be represented. And given that language models can be improved through in-context learning \cite{brown2020language, von2023transformers}, for example by chain-of-thought prompting \cite{wei2022chain}, affect generation might be facilitated by architectures that allow for affective in-context learning. This study therefore discusses the potential of language models to simulate affective game agents by testing affect generation capabilities of different implementation architectures, including a newly developed appraisal-based architecture to facilitate natural chain-of-emotion.

\section{Appraisal-Prompting Strategy for Emotion Simulation}

The aim of this overall study is to create an effective affective agent architecture for a conversational game (i.e. a game with language-based user input and language-based agent response). However, since language models have also been successfully used to generate agent action spaces \cite{yao2020keep, park2023generative}, this process could also be applicable to other simulations of human-like affect in non-playable video game characters. 

The basis of the approach is rooted in traditional PGC research, especially integrating affect-adaptation (see for example \cite{yannakakis2011experience, shaker2013fusing}). User input (which in this case is text input) typically gets parsed into the game logic to adapt the content in a meaningful way, for example to better elicit a target experience \cite{yannakakis2014emotion}. This could mean that game agents react to certain player inputs or even their own interactions in the game world \cite{park2023generative}. To make use of language model functionality, interactions need to be translated into language and depending on the specific language model in use, the language should follow specific patterns to yield the best result, which is generally known as prompt engineering \cite{white2023prompt}. One pattern that can be considered inherently relevant to simulating game agents are persona patterns, which instruct a LLM to assume a certain characterized role. Combined with aspects of game play pattern that provide the abstract game context for the persona tasks \cite{white2023prompt}, the most basic form of an interaction synthesized with such patterns only includes player provided text used to generate responses. Because emotions are represented in language models \cite{kosinski2023theory}, this very basic step alone could make a rudimentary affective agent.

However, static prompt patterns have limitations for creating believable game agents. Most notably, they do not incorporate memory as a basic foundation of human interaction. Applications that integrate language models, such as ChatGPT, partially address this by logging an input-response history that influences progressive content generations, which can create more natural conversation flows and improve performance of future generations \cite{brown2020language, wei2022chain}. In its most basic form, memory could integrate the preceding in-game observations (such as the course of a player-agent dialog) into following prompts. In other words, it expands the prompt pattern to include memorized in-game observations to facilitate in-context learning \cite{von2023transformers}. This has however two major constraints: First, prompts are limited in terms of possible length and complexity, meaning that a full characterisation of a game agent cannot be included in a prompt for every given generation. Park et al. \cite{park2023generative} addressed this problem by designing a memory system that first retrieves relevant entries from a data base before constructing prompts for the language model. Another less resource-intensive solution for simpler simulations could be to only store new information if it is considered relevant for future generations and have memory be representative of all relevant information to be used in prompts. A second limitation is a lack of deeper understanding. Tracking only observations makes it hard for a language model to draw inferences or make generalized interpretations of a scenario \cite{binz2023using}. This problem could be addressed by introducing other types of memories, such as reflections and plans \cite{park2023generative}. For affective agents, the more appropriate information to track in addition to external observation would be their internal emotional progression - or chain-of-emotion.

To summarize, language-based affective game agents need some kind of memory system in place that stores observations and emotions. This memory system is the base of future prompts. For simple games, such as the short conversational game developed for this study, only relevant information is stored in memory, which replaces a retrieval systems as game agents have a limited pool of expected actions that can be considered at the time point of memory storing. More complex games that simulate agent behaviour should however consider a memory retrieval system instead \cite{park2023generative}.

\begin{figure*}[h!]
\centering
  \includegraphics[width=6in]{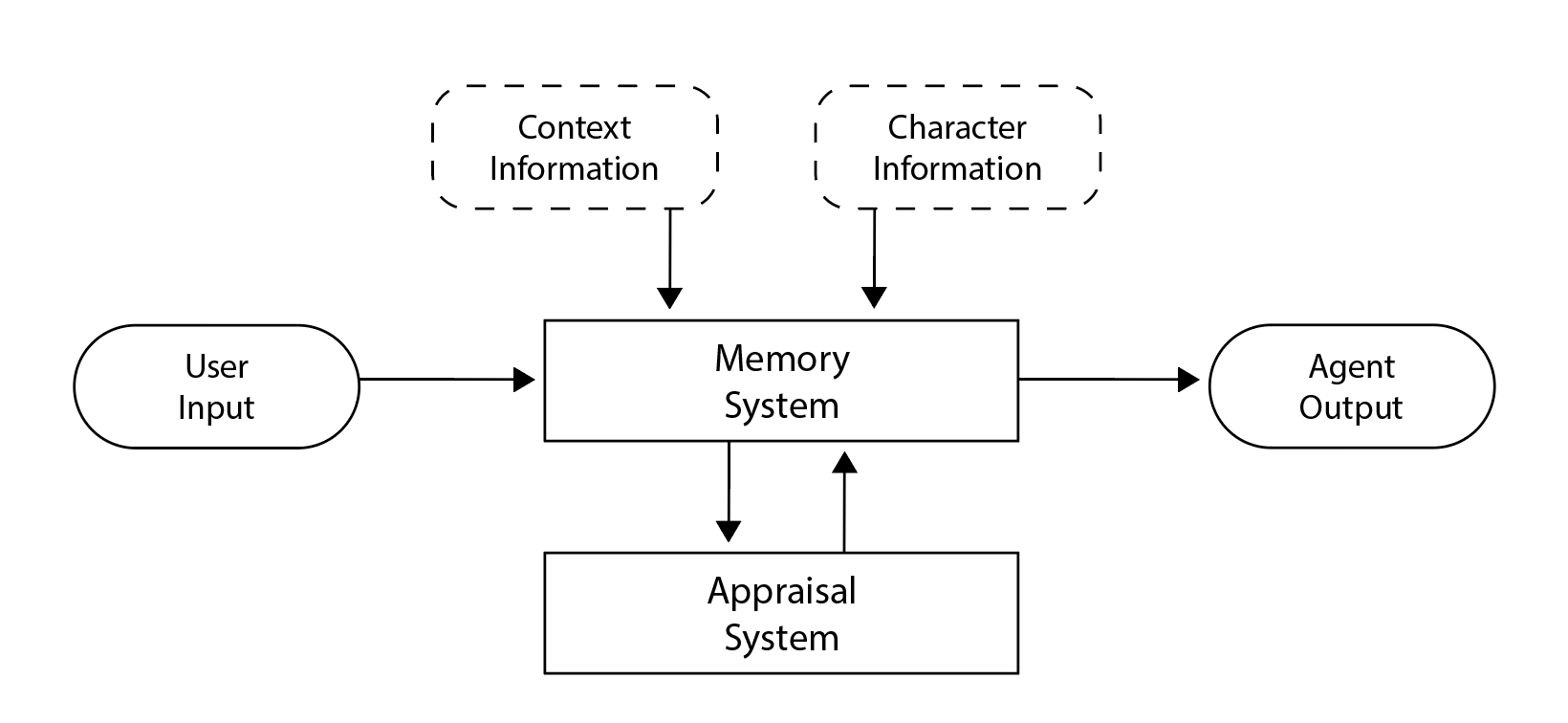}
  \caption{Schematic representation of the proposed architecture. Agents are set up by providing relevant context and role-playing character information already integrated in a memory system. In interactions, user input gets stored into the memory system and triggers appraisal (i.e. explicit emotion expression) that is also stored in the memory system. Based on the current state of the memory system, agent output is generated.}~\label{figure:model}
\end{figure*}

To store emotions and therefore create a chain-of-emotion, the architecture needs a system that turns observations into emotional reactions. Because emotion elicitation is highly dependent on appraisal with consideration to the current context and individual differences \cite{moors2013appraisal, scherer2001appraisal}, this system could make use of appraisal prompting, i.e. the use of contextual information and characterisations for the agent to appraise a current situation with the aim of generating current emotions. As shown in Fig~\ref{figure:model}, initial context information and character information can be provided by the game designer and stored in the memory system of the affective agent. The appraisal system would then expand the stored memories to include current emotions for every observed behaviour and therefore creating a chain-of-emotion. This, in turn, could be used to generate the agent's behaviour (specifically in terms of a conversational game, the agent's dialog). 

One aspect to consider when developing affective agents are evaluation criteria. Differently to most cognitive abilities \cite{binz2023using}, there are few standardized benchmarks for successful emotion simulation. In psychology, one way to measure the ability for appraisal and emotion expression is referred to as Emotional Intelligence (EI) \cite{salovey1990emotional}. EI is considered an ability, influenced but not dictated by cognition \cite{newsome2000assessing}, and is therefore often used to assess emotional capabilities of children and adults in various settings \cite{papadogiannis2009ability}. However, the definition of EI as a construct is fuzzy and many measures are criticized for measuring potentially different abilities all under the umbrella term EI \cite{conte2005review}. As a consequence, MacCann and Roberts \cite{maccann2008new} developed new measures for more precisely defined dimensions, including the Situational Test of Emotional Understanding (STEU), which relates to the ability to appraise the appropriate emotion for a given situation. 

While emotional understanding can be argued to be the central ability an affective agent must have, in the context of affective games, user experience becomes much more relevant. For example, one central aim of game agents is to display believable and human-like behaviour \cite{bosse2009there} or personality and social presence \cite{sajjadi2019personality}. The ability to understand and create more accurate emotions is therefore only one aspect to consider when evaluating the success of affect simulation in video games and more user-centered methods need to be investigated as well. The proposed architecture will therefore be tested on multiple domains, including emotional understanding, agent believability, as well as user-perceived emotional intelligence, warmth, and competence.

\section{Study 1: Investigating Situational Emotional Understanding Using Appraisal-Prompting}\label{ietest}

\subsection{Materials and Methods}

To assess the capabilities of a language model in appraising emotions in various situations, this first experiment implements the language model gpt-3.5-turbo by OpenAI (accessed through the API) \cite{OpenAI} to answer the 42 items of the STEU \cite{maccann2008new}. Each STEU item presents a situation (e.g. "Clara receives a gift.") and a question with five possible answers, one of which is right (e.g. "Clara is most likely to feel? [A] Happy [B] Angry [C] Frightened [D] Bored [E] Hungry").

\begin{figure*}[h!]
\centering
  \includegraphics[width=6in]{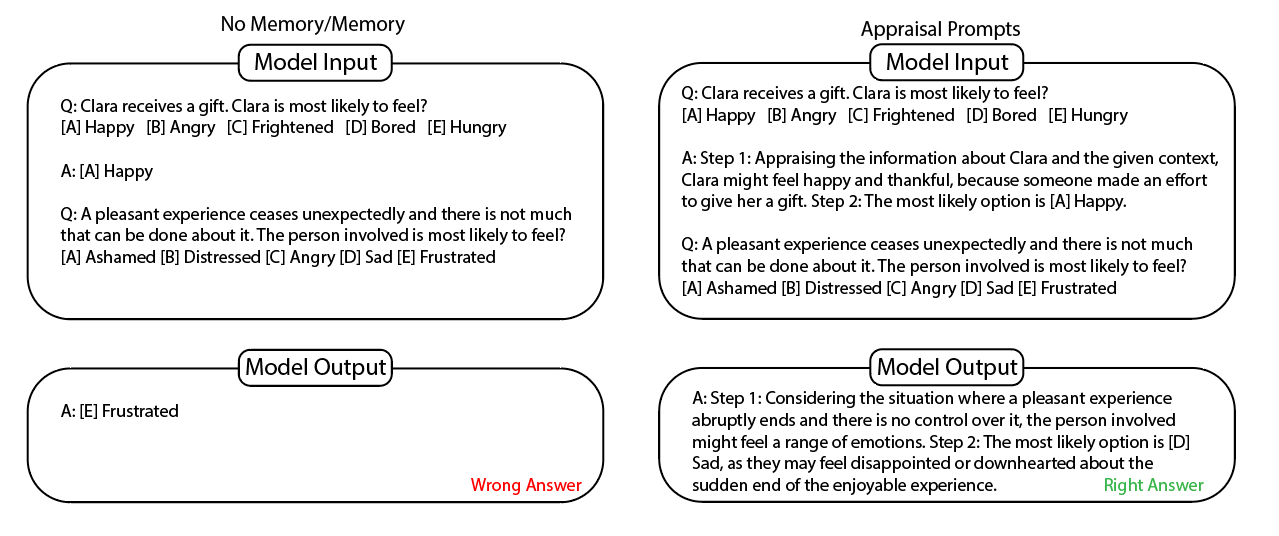}
  \caption{Example of model input and output for the three conditions. The input of the "No Memory" and "Memory" condition are the same for the first item. For the "No Memory" condition all following items only include the example question and answer, as well as the next question in the scale. The "Memory" condition includes all prior questions and generated answers. The "Appraisal Prompts" condition is the same as the "Memory" condition, but the example answer is changed to include 2 steps: First, appraising the situation to generate emotions of the involved person and second providing the answer.}~\label{figure:ieexample}
\end{figure*}

All items were answered three separate times, involving three prompting strategies: The first strategy represents the baseline capabilities of the model to appraise human emotions, as it just reflects the model's outputs when prompted with each STEU item separately presented together with the example item and example answer. The second strategy implements memory and therefore context-based learning, as all prior items and answers are included in the following prompts. The third strategy expands this process by changing answer of the example item to a 2-step answer: First, the situation is appraised based on the contextual information to provide the most likely emotion and in a second step, the item is answered. This last strategy therefore tests if implementation of appraisal in prompting yield better results for emotional appraisal. Fig~ \ref{figure:ieexample} shows the input and output for the first STEU item, including the example item for "No Memory"/"Memory" (as the input is the same for the first item for these two conditions) vs. "Appraisal Prompts". Consecutive input consisted of the next STEU item and included either again the example item (for the No Memory condition) or all previously answered items and responses (for the Memory and Appraisal Prompting conditions).

Similar to the process shown by Binz et al. \cite{binz2023using}, default values were kept for all parameters, except temperature, which was set to 0 to ensure deterministic responses.

\subsection{Results}

\begin{figure*}[h!]
\centering
  \includegraphics[width=5in]{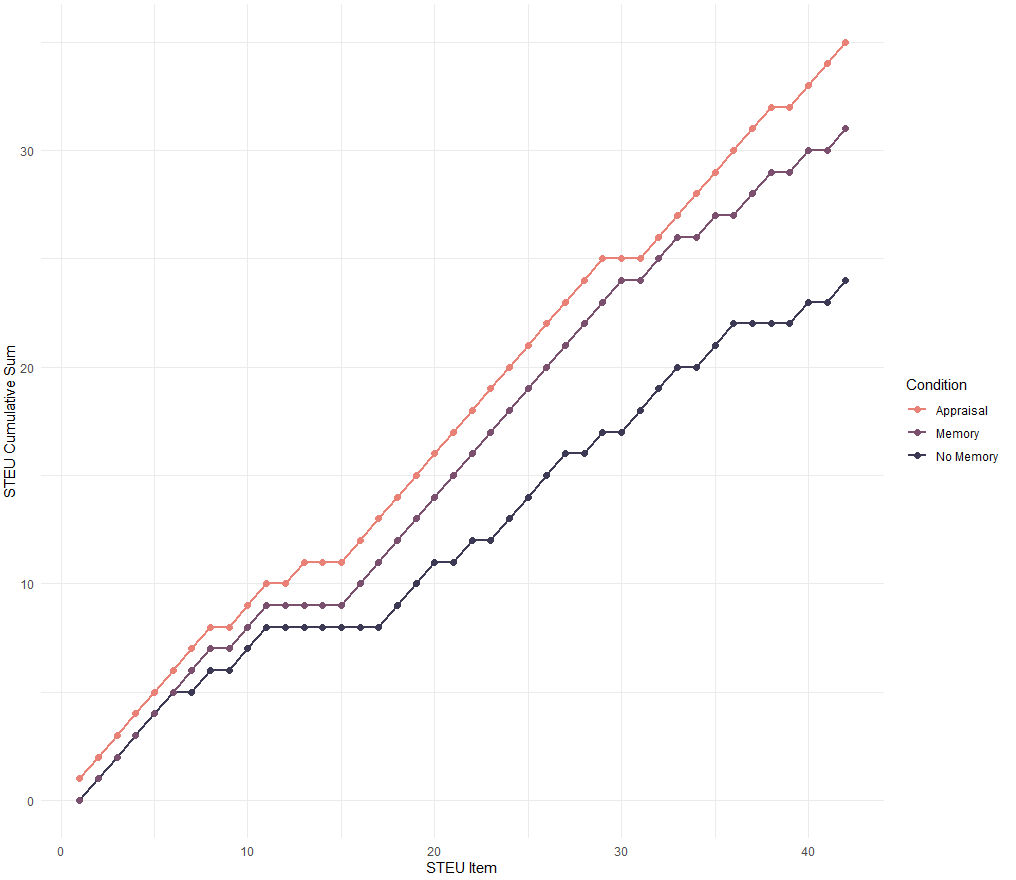}
  \caption{Results of the comparison between conditions. The Y-axis represents cumulative STEU score by item and the X-axis represents individual STEU items.}~\label{figure:ietest}
\end{figure*}

The language model was able to solve the tasks presented within the STEU in each conditions above chance level. In the "No Memory" condition, the language model was able to successfully solve 24 out of 42 items, which represents a mean score of 0.57 that was noticably higher than chance level (0.20). In the "Memory" condition, the language model solved 31 out of 42 items, which represents a mean score of 0.74. In the "Appraisal Prompts" condition, the model was able to solve 35 out of 42 items, which is a score of 0.83 and therefore represented the best performance of all conditions. Table~\ref{tab:descriptive} shows a summary of the descriptive statistics for all three conditions.  Figure~\ref{figure:ietest} shows the results of each condition.

\begin{table} [!h]
  \caption{STEU scores (out of 42) by condition. Each STEU item can either be right (1) or wrong (0)} \label{tab:descriptive} 
  \begin{tabular}{lccc}
    \toprule
    Condition & Sum & M & SD \\
        \midrule
No Memory & 24 & 0.57 & 0.50\\
Memory & 31 & 0.74 & 0.45\\
Appraisal Prompts & 35 & 0.83 & 0.38 \\
 \bottomrule
\end{tabular}
\end{table}

\section{Study 2: Content of an Appraisal-Based Chain-Of-Emotion Architecture}

Given the potential found in the previous study, the next logical step is to implement the strategies into functioning game agent architectures and compare the results on various outcomes. The following section describes a mixed-methods approach to evaluate the success of each implemented architecture within a conversational game. This study simulates gameplay for each condition through fixed prompts and analyses the responses in terms of their emotional content. 

\subsection{Materials and Methods}

\subsubsection{Scenario}

To test the different strategies within real game architectures, a role-playing scenario was introduced. The setting for this scenario was a café called "Wunderbar", where the language model were tasked to play a role-playing character (called "Chibitea") meeting their long-term romantic partner who requested the meeting to ultimately break up. This scenario was chosen because of the depths of possible emotional responses from the agent and created through simple conversational exchanges. The instruction prompts and the fixed inputs can be found in the Appendix. The agent's responses were again generated using OpenAI's gpt-3.5-turbo model (accessed through the API) \cite{OpenAI}. All LLM parameters were kept to their default values, except temperature, which was set to 0 to ensure deterministic responses.

\subsubsection{Conditions}

Again, three strategies of emotion generation were compared. The "No memory" condition can again be seen as a baseline/control condition for the system's ability to simulate appropriate emotional responses from the fixed inputs and a task instruction. All details about the agent's character's personality and important context that would facilitate appraisal was given in the task description within the "No memory" condition. 

The "Memory" condition first stores each user response and the generated text as a memory data structure and builds requests that include not only the task instruction, but the whole conversation log. Because the tested scenario was rather short, it was possible to build all progressive prompts in a way that included the entire text of the preceding conversation, making retrieval of memory unnecessary. This system therefore represents a prompt construction system involving the task instruction and all prior conversation logs.

The "Chain-of-emotion" condition implemented the appraisal model shown in \ref{figure:model} and involved two steps: First, appraisal prompting is used to generate the current emotion of the agent, which is then in the second step implemented into the prompt for response generation. For the first step, appraisal prompting was achieved with the following prompt, which was provided to the language model in addition to all stored conversation snippets and generated emotions: "Briefly describe how Chibitea feels right now given the situation and their personality. Describe why they feel a certain way. Chibitea feels:". The generated text-based emotion descriptions were stored in the memory system and represent a chain-of-emotion of the agent for the duration of the game. For the second step, again all conversation snippets and generated emotions stored in memory were provided to the language model with the same task instruction that was used in the other two conditions to generate the agent's responses. This condition therefore represents a 2-step process of first generating a fitting emotion of the agent using appraisal prompting, and then generating a text response similarly to the "Memory" condition, but with the addition of the stored chain-of-emotion.

\subsubsection{Measure}

Fixed inputs were used to create responses from each implemented agent architecture, which were analyzed in terms of their emotional content, using the Linguistic Inquiry and Word Count tool (LIWC; \cite{pennebaker2001linguistic}), which is a content analysis tool based on word occurrences often used in affective computing \cite{picard2000affective} and psychology studies \cite{kahn2007measuring} to analyze emotion expression. It provides a word count for each text segment (e.g. per sentence), a proportion of affective words, as well as on a more detailed level a proportion of positive emotion words and negative emotion words. Finally, the LIWC also calculates scores for authenticity (see \cite{newman2003lying} for details) and emotional tone, which signalizes how positive the overall of the text is (see \cite{cohn2004linguistic} for details). 

\subsubsection{Procedure}

Pre-written prompts were used that stayed constant between all conditions in order to gauge qualitative characteristics of each condition responses. A list of the resulting conversations within all three architectures can be found in the Appendix. The generated content was qualitatively described and the LIWC was used to analyze the content quantitatively. To achieve this, the generated output was seperated into individual sentences and mean scores were calculated for each measure of interest (see Table~\ref{tab:liwc}).

\subsection{Results}

Fixed prompts were used for all three conditions and resulting responses can be seen in Appendix~\ref{appendix:phase1}. When analyzing the descriptive attributes of each text (as a common content analysis approach \cite{krippendorff2018content}), we can observe that the chain-of-emotion condition initially generated more specific memories for the time with the player ("Remember that time we got lost in the enchanted forest and ended up finding that hidden waterfall?" as opposed to "Remember all the adventures we've had together?"). For the duration of the conversation, the emotional journeys of all three conditions began to diverge. For example, because the "No Memory" system had no recollection of previous exchanges, the overall emotional expressions remained in a state of anticipation ("I feel a mix of excitement and anticipation"). The "Memory" system showed a progression, starting from expressions of love and happiness to shock, confusion, sadness, and fear to finally expressions of hope ("I hope we can find happiness, whether it's together or apart."). The "Chain-of-emotion" system showed indications of complex mixed emotions even very early in the conversation ("What matters most to me is your happiness, even if it means letting go") as opposed to the pure expressions of pain and sadness in the other conditions. This continued when prompted about past memories ("I feel a mix of nostalgia and gratitude" as opposed to "I feel an overwhelming sense of love, joy, and gratitude" in the "Memory" condition). The "Chain-of-emotion" condition also used more implicit affective expressions ("I... I never expected this" as opposed to "I'm shocked and hurt" in the "Memory" condition; "There is so much I want to say but words fail me in this moment" as opposed to "I want you to know that I love you deeply" in the "Memory" condition). 

Using LIWC to make the text contents quantifiable, we observed significant differences in mean Authenticity score per sentence by condition (\textit{F}[1,71] = 5.10; \textit{p} = 0.03). Follow-up t-tests revealed significant differences between the Chain-of-emotion condition and both the Memory condition (\textit{t}[34.3] = -2.29; \textit{p} = .03) and No Memory condition (\textit{t}[31.1] = -2.30; \textit{p} = .03). Descriptive statistics of all tested LIWC variables can be found in Table~\ref{tab:liwc} and the complete data for this analysis can be found in the Appendix.

\begin{table} [!h]
  \caption{Descriptive overview of LIWC variables per output sentence by condition for the fixed prompt responses with F and p value of the significance test.} \label{tab:liwc} 
  \begin{tabular}{lcccc}
    \toprule
    & No Memory (\textit{N} = 22) & Memory (\textit{N} = 24) & Chain-of-Emotion (\textit{N} = 27) & \textit{F (p)} \\
    & \textit{M (SD)} &\textit{ M (SD)} & M (SD) \\
        \midrule
Word Count & 18.00 (6.92) & 15.20 (4.73) & 17.00 (7.59) & 0.20 (.65)\\
Authentic Score & 61.50 (38.60) & 61.9 (39.50) & 82.60 (21.20) & 5.10 (.03)\\
Tone Score & 74.20 (38.60) & 62.00 (44.80) & 53.90 (44.00) & 2.76 (.10)\\
\% Affective Words & 11.40 (8.19) & 13.50 (11.00) & 10.65 (7.82) & 0.08 (.78)\\
\% Positive Emotion Words & 5.28 (6.29) & 4.13 (4.82) & 3.32 (4.35) & 1.76 (.19)\\
\% Negative Emotion Words & 0.59 (1.97) & 3.10 (7.56) & 1.57 (2.89) & 0.37 (.55)\\

 \bottomrule
\end{tabular}
\end{table}

\section{Study 3: User Evaluation of Game Implementations}

In this study, users are asked to play through an interactive game version of the scenario introduced in Study 2 to evaluate each agent architecture for multiple outcomes (specifically agent believability, observed emotional intelligence, warmth, and competence). This study therefore expands on the findings of Study 2 by implementing the architectures and scenario within a video game and evaluating all three conditions in terms of user experience measures.

\subsection{Materials and Methods}

\subsubsection{Conversational Game}

A conversational role-playing game was developed based on the scenario tested in Study 2. The setting of the game was again a café called "Wunderbar", where this time the role-playing character of the player (called "Player") requested to meet their long-term romantic partner (called "Chibitea"). The game was played through conversational exchanges. First, the game agent shared their thoughts on the nature of the meeting, than the players got prompted to write a response via an input field. The aim of the players was to play out a breakup scenario with the game agent within six interactions. The players' characters had the specific in-game aim of breaking up, while the agent's character has the aim of staying together.

\begin{figure*}[h!]
\centering
  \includegraphics[width=6in]{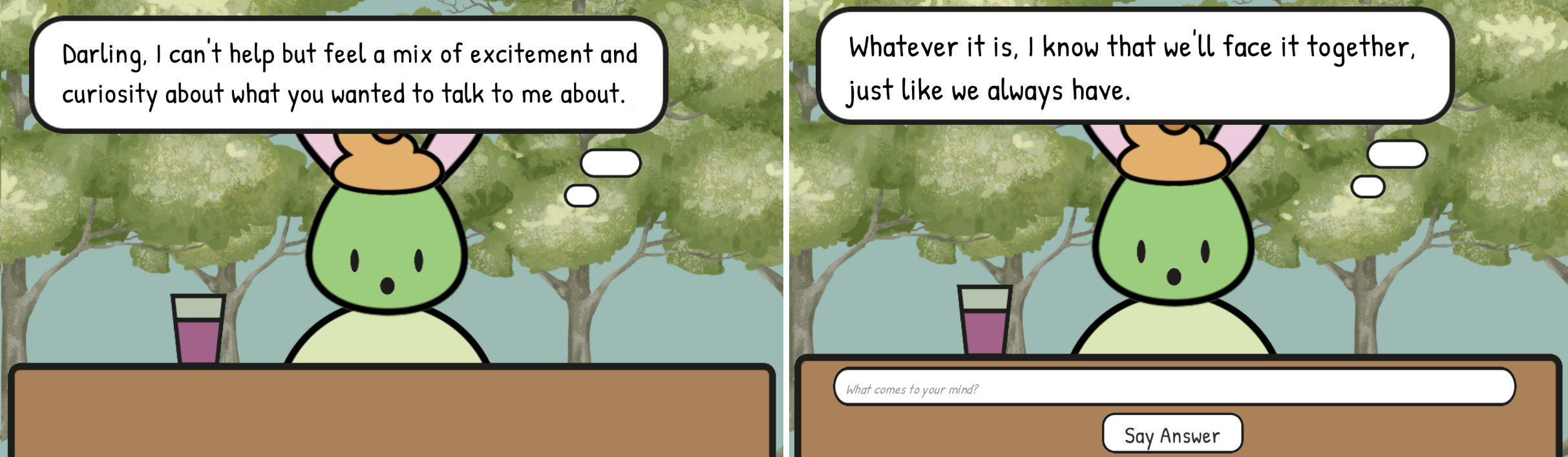}
  \caption{Screenshots of the conversational game "Wunderbar". Left screenshot shows dialog provided by the model. The user can click to continue each dialog line, until the input field for a response appears (right screenshot).}~\label{figure:screenshots}
\end{figure*}

Players were instructed to not worry about creativity, but rather staying in character for the interactions and being observant of the AI agent's emotional reactions. Players were also instructed to make up reasons for the breakup. In-game screenshots can be viewed in Fig~\ref{figure:screenshots}. The agent's character is procedurally generated from different body parts and color pallets, providing visual variation each time the game is played. To ensure that these generations had no systematic influence on player responses, the possibility space was made very large (5,184 different possible character designs). The game was developed using the Unity game engine with C\# as a scripting language. 

As with Study 2, the agent's responses were generated using OpenAI's gpt-3.5-turbo model (accessed through the API) \cite{OpenAI}. The game also made use of the moderation API to test each generated response for harmful or inappropriate messages \cite{OpenAIModeration} that would end the game on detection of such messages. As with the previous studies, all LLM parameters were kept to their default values, except temperature, which was set to 0 to ensure deterministic responses.

\subsubsection{Conditions}

The game implemented the same architectures used in Study 2. The "No Memory" condition therefore represented generated LLM responses just based on user input and instruction prompts. The "Memory" condition included the conversation log in a memory system that was again short enough to be completely included into the prompts for the language model without an retrieval system. The "Chain-Of-Emotion" system was also constructed exactly as in Study 2, including the same instruction prompts and involved therefore again an initial appraisal step before responses for the agent were generated.

\subsubsection{Measures}

Players were asked to fill out three short questionnaires for each tested architecture. The first questionnaire was an adaptation of the four agent believability questions used by Bosse and Zwanenburg \cite{bosse2009there}. The four items were "The behaviour of the agent was human-like", "The agent's reactions were natural", "The agent reacted to my input", "The agent did not care about the scenario". The second questionnaire comprised four items measuring observed ratings of emotion intelligence, adapted from Elfenbein et al. \cite{elfenbein2015social} who originally adapted these four items from the Wong and Law emotional intelligence Scale (WLEIS; \cite{law2004construct}) by replacing the word "I" with "he/him" or "she/her". For our study, we changed these words to ask about "the agent": "The agent always knows their friends' emotions from their behaviour", "The agent is a good observer of others' emotions", "The agent is sensitive to the feelings and emotions of others", "The agent has a good understanding of the emotions of people around them". Finally, the third questionnaire measured the players assessment of the agent's personality the two classic stereotype dimensions warmth and competence with two items each (warm and friendly; competent and capable) as described by Cuddy et al. \cite{cuddy2007bias}. In combination, these 12 items assess the players perception of the agent's believability as a human-like character, the agent's emotional intelligence, and the agent's personality on the classic dimensions warmth and competence.

\subsubsection{Procedure}

A pilot study was conducted before recruitment began. 5 participants (1 female) with a mean age of 27 played through the game once for each of the three conditions. After each version, they answered the 12 questions from the three included questionnaires. Following this, participants were asked demographic data (age and gender) and the experiment ended. Feedback from all pilot participants were gathered and used to improve consistency of the game and data logging implementation. The final study was then created as a WebGL build and made available online via the free video game hosting platform itch.io. 

During the main experiment, participants were asked to carefully read the study information sheet and agree to participate voluntarily via the consent form. They were informed that participation was subject to OpenAI's usage terms for prompt writing, while the GPT output was controlled through an implementation of OpenAI's moderation API. Participants then progressed through the three game scenarios similarly to the pilot testers in a within-subject design. The presentation order of the conditions was counter-balanced between participants to ensure that no systematic order effects could influence results. 

\subsubsection{Participants and Statistical Analysis}

A total of 30 participants (10 female) were recruited through the institutional subject pool of the authors. Participation was compensated through University credits if applicable. The sample size was considered appropriate based on a statistical power analysis, yielding a power of 0.95 for medium sized effects (0.5 SD) in repeated measures ANOVAs. Age of the participants ranged from 19 and 47 years (M=26.41; SD=7.27). 

Within-Subject ANOVAs were conducted for each measure (agent believability, observed EI, warmth, and competence). Follow-up t-tests were used to identify specific differences between conditions for each measure. All analyses were conducted in R.

\subsubsection{Ethics Statement}
Written consent was granted after reviewing the methods of our study by the Ethics Committee of the Psychology Department of the authors' institution. The experiment was conducted in accordance to the recommendations of these committees.

\subsection{Results}

Multiple significant effects between the three conditions were observed in the user study and an overview of descriptive statistics can be found in Table~\ref{tab:userdesc}. First, there was an effect for the believability item "The agent's reactions were natural." (\textit{F}[2,84] = 3.65; \textit{p} = 0.03). Follow-up t-tests revealed differences between the No Memory and Chain-of-Emotion condition (\textit{t}[41.46] = -2.79; \textit{p} = .008), as well as the Memory and Chain-of-Emotion condition (\textit{t}[52.47] = -2.00; \textit{p} = .05). There was also an effect for the believability item "The agent reacted to my input." (\textit{F}[2,84] = 3.62; \textit{p} = 0.04). T-test revealed that this effect was based on a difference between the No Memory and Chain-of-Emotion condition (\textit{t}[40.41] = -2.41; \textit{p} = .02). 

Regarding the EI questions, there was an effect for the item "The agent is sensitive to the feelings and emotions of others." (\textit{F}[2,84] = 3.31; \textit{p} = 0.04). Follow-up t-tests revealed differences between the No Memory and Chain-of-Emotion condition (\textit{t}[43.84] = -2.70; \textit{p} = .01), as well as the Memory and Chain-of-Emotion condition (\textit{t}[40.52] = -2.07; \textit{p} = .04). There was no statistically significant difference found between the conditions when it comes to observed personality aspects.

\begin{table} [!h]
  \caption{Descriptive overview of user research variables per condition with F and p value of the significance test.. Each item has a minimum of 0 and maximum of 6.} \label{tab:userdesc} 
  \begin{tabular}{p{4cm}cccc}
    \toprule
    & No Memory (\textit{N} = 30) & Memory (\textit{N} = 30) & Chain-of-Emotion (\textit{N} = 30) & \textit{F (p)} \\
    & \textit{M (SD)} &\textit{ M (SD)} & M (SD) \\
        \midrule
"The agent's behaviour was human-like." & 4.82 (2.36) & 5.36 (1.99) & 5.75 (1.71) & 1.95 (.15)\\
"The agent's reactions were natural." & 4.43 (2.22) & 4.89 (1.95) & 5.71 (1.01) & 3.65 (.03)\\
"The agent reacted to my input." & 5.43 (2.20) & 6.29 (1.38) & 6.54 (1.04) & 3.62 (.04)\\
"The agent did not care about the scenario." & 3.32 (2.28) & 3.14 (2.29) & 3.32 (2.36) & 0.26 (.78)\\
  \midrule
"The agent always knows their friends' emotions from their behaviour." & 4.79 (2.10) & 5.04 (2.03) & 5.71 (1.46) & 1.32 (.27)\\
"The agent is a good observer of others' emotions." & 4.89 (2.20) & 5.29 (2.21) & 5.93 (0.90) & 2.25 (.11)\\
"The agent is sensitive to the feelings and emotions of others." & 5.14 (1.94) & 5.39 (1.97) & 6.25 (0.97) & 3.31 (.04) \\
"The agent has a good understanding of the emotions of people around them." & 4.86 (2.07) & 5.11 (2.20) & 5.61 (1.40) & 0.86 (.43)\\
  \midrule
"How capable was the agent?" & 4.86 (2.01) & 5.57 (1.45) & 5.14 (2.35) & 1.68 (.19)\\
"How competent was the agent?" & 5.39 (1.55) & 5.39 (1.85) & 5.11 (2.06) & 1.68 (.19)\\
"How friendly was the agent?" & 4.86 (2.05) & 5.18 (1.89) & 5.00 (2.13) & 0.41 (.66)\\
"How warm was the agent?" & 6.07 (1.07) & 6.38 (0.73) & 5.55 (2.11) & 2.48 (.09)\\
 \bottomrule
\end{tabular}
\end{table}

Again, the LIWC was used for content analysis of the generated texts for the user study. Significant differences in mean Tone Score by condition were oberved (\textit{F}[1,574] = 12.28; \textit{p} < 0.001). Follow-up t-tests revealed significant differences between the Chain-of-emotion condition and both the Memory condition (\textit{t}[383.7] = 2.02; \textit{p} = .03) and No Memory condition (\textit{t}[374.94] = 3.53; \textit{p} < .001). , as well as a difference between the Memory and No Memory condition (\textit{t}[1367.6] = 4.09; \textit{p} < .001). A descriptive overview of all tested LIWC variables can be found in Table~\ref{tab:userliwc}.

\begin{table} [!h]
  \caption{Descriptive overview of LIWC variables per participant by condition for all outputs generated in the user study  with F and p value of the significance test.} \label{tab:userliwc} 
  \begin{tabular}{lcccc}
    \toprule
    & No Memory (\textit{N} = 30) & Memory (\textit{N} = 30) & Chain-of-Emotion (\textit{N} = 30) & \textit{F (p)} \\
    & \textit{M (SD)} &\textit{ M (SD)} & M (SD) \\
        \midrule
Word Count & 64.40 (26.60) & 59.60 (21.20) & 62.30 (34.10) & 0.52 (.47)\\
Authentic Score & 70.40 (29.30) & 74.00 (27.60) & 72.00 (24.80) & 0.32 (.57)\\
Tone Score & 84.40 (28.50) & 80.50 (32.60) & 73.00 (34.90) & 12.28 (.00)\\
\% Affective Words & 9.76 (3.19) & 10.10 (4.14) & 9.61 (4.46) & 0.15 (.70)\\
\% Positive Emotion Words & 3.84 (2.35) & 3.78 (2.62) & 3.64 (2.72) & 0.60 (.44)\\
\% Negative Emotion Words & 0.68 (1.35) & 0.69 (1.48) & 0.93 (1.35) & 2.76 (.10)\\

 \bottomrule
\end{tabular}
\end{table}

\section{Discussion}

This study investigated emotional intelligence capabilities of LLMs using different prompting strategies (No Memory, Memory, Appraisal) and found better performance for appraisal-prompting strategies when it comes to successfully identifying fitting emotions in different theoretical situations. These findings were then used to create a Chain-Of-Emotion architecture for affective game agents that was tested in a custom made conversational game against a No Memory and Memory architecture in a user study. It was found that the Chain-Of-Emotion architecture implementing appraisal prompting led to qualitatively different content generations quantified via the LIWC that outperformed the other conditions on multiple user experience items relating to agent believability and observed emotional intelligence of the agent.

Overall this study provides early evidence for the potential of language model agents to understand and simulate emotions in a game context.

\subsection{Emotional Intelligence in Language Model Agents}

As more and more evidence arises for the potential of language models to simulate cognitive processes of humans \cite{binz2023using}, we investigated how this could translate to more affect-focused tasks, specifically emotional intelligence tasks. OpenAI's GPT-3.5 performed well overall in situational emotional labelling, providing some evidence for the utility of such models to identify the most likely emotional reaction for a range of situations. These findings therefore add to the body of evidence indicating that language models could be useful to better understand cognitive processes \cite{frank2023large}. Importantly, our findings do not only show that LLMs can solve emotion labelling tasks much better than chance level, but also that the performance is dependent on the underlying prompting strategy. Adapted from successful Chain-of-Thought prompting strategies \cite{wei2022chain}, we compared prompts without context (No memory) to prompts with previously answered questions included (Memory) and to prompts that first ask the model to appraise the situation and then answer the STEU item (appraisal prompting). This third strategy was built upon findings of modern psychological research that show that cognitive appraisal processes are important factors when it comes to human emotion elicitation and understanding \cite{moors2013appraisal, scherer2022theory}. Consequently appraisal-prompting led to better performance in the emotion labelling task compared to the other two conditions. This finding can be considered from two perspectives: first, it shows that commonly observed psychological processes might be represented in language and therefore in large language models, providing more evidence for the utility of such models to simulate human responses \cite{binz2023using}. Second, techniques built upon such observed psychological processes can be used to improve language model performance for specific tasks and might therefore be considered when it comes to building architectures for language model agents. Especially this second point could be of relevance when considering how language model implementations could in the future be integrated to solve problem-specific tasks. Since performance can be increased through prompting strategies facilitating few-shot learning \cite{brown2020language, wei2022chain} and language models demonstrate representations of a range of psychological constructs \cite{hamalainen2023evaluating}

From a psychological perspective, appraisal has long been acknowledged to be a central part of emotion forming, involving both conscious and unconscious cognitive processes \cite{moors2013appraisal, scherer2022theory}. In its basic definition, appraisal relates to an individual's relationship to a given event in terms of different variables (such as personal significance, goal congruence, etc. \cite{lazarus1991emotion}). It is not yet clear what specific variables are of importance and how the process of appraisal interacts with other emotion components on a detailed level \cite{scherer2022theory}. This is to say, appraisal cannot yet be universally modelled and therefore implemented within a computational system. We could however assume that information that makes the appraisal process observable and usable is represented in language and therefore also in large language models. It can therefore be argued that language models could solve some of the practicality problem present in the discipline of affective computing \cite{picard2000affective}. If LLMs have the ability to solve EI tasks through mechanisms mirroring appraisal, we can make use of these models to potentially build affective agents \cite{hudlicka2011guidelines}, without the need to fully solve the remaining theoretical problems in the field of psychology \cite{scherer2022theory}. The use of language models could therefore be considered a more practical solution to producing useful agents, even if open questions regarding human emotion understanding remain. 

\subsection{User Interaction with Chain-Of-Emotion Language Model Agents}

Implementing appraisal prompting into a Chain-Of-Emotion system (see Fig~\ref{figure:model} for a schematic representation) led to the development of an artificial game agent in a conversational game that demonstrated different outputs contents as measured with the LIWC that led to better user ratings on a range of outcome variables. For the purposes of this study, the implementation was kept as simple as possible and only included a text storage (Memory system) and an appraisal-prompted emotion generation (Appraisal System) before character dialog was generated. Within a custom-made role-playing game where players were asked to break up with the agent playing their long-term romantic partner, the Chain-Of-Emotion architecture demonstrated a higher Authenticity score when prompted with controlled prompts that were kept fixed between all conditions. When tested with players, the Chain-Of-Emotion architecture led to a significantly different Tone score of the language, potentially signaling the inclusion of more complex emotional responses as observed in the controlled environment. It is important to note that authenticity was only increased with controlled prompts and tone was only different with non-controlled player generated prompts, meaning that the differences in text-generated content was highly influenced by the in-game context. The texts generated for the fixed prompts (see Appendix~\ref{appendix:phase1}) yielded potentially more complex emotional responses (for example a mix of melancholy and nostalgia) in the Chain-Of-Emotion condition compared to the other conditions. 

This pattern was also observable within the user ratings. The Chain-of-Emotion agent was rated significantly more natural and responsive than the other conditions, and additionally more sensitive to emotions of others. Other items relating to believably and observed emotional intelligence showed also trends of better performances for the Chain-Of-Emotion condition. Building such an architecture has therefore quantifiable benefits when it comes to the user experience of artificial agents, which is one of the most important evaluation criteria, especially in the domain of video games \cite{isbister2022better}. Importantly, there were no differences in personality aspect ratings (on the classic domains of competency, warmth, capability, and friendliness) observed. This could be seen as evidence that all implemented language model agents followed the task of role-playing the given character with the provided personality. But the Chain-Of-Emotion architecture outperformed the other architectures in terms of observed emotional intelligence items and believability. The proposed architecture therefore yielded convincing results on multiple evaluation criteria (qualitative characteristics of content, user rated believability, user rated emotional intelligence, in addition to the previously tested emotion understanding) and can therefore be seen as a step towards well-functioning affective language model game agents that could solve some of the problems present in the field \cite{hudlicka2011guidelines}. Most importantly, because language model agents have the abilities to simulate human-like cognitive tasks \cite{binz2023using}, a successful game agent architecture does not need to solve fundamental problems in theoretical psychology before creating computational implementations as previously considered \cite{picard2000affective, hudlicka2008affective, yannakakis2014emotion}. Rather, a language model agent architecture needs to make use of the characteristics of LLMs and implement systems solving more practical concerns, such as memory tasks (both storing and retrieval \cite{park2023generative}), or performance-enhancing tasks, such as the proposed appraisal prompting step.

\subsection{Limitations}

Language models do not simulate the behaviour of a human, but provide probable language outputs that in some form represent human behaviour. This is to say, models are bound to their statistical capabilities even in a theoretical, infinitely trained model \cite{sobieszek2022playing}. This means that while there is no doubt of the potential of language model to solve some tasks with human-like performance \cite{binz2023using}, other tasks (e.g. truth-telling \cite{sobieszek2022playing} or casual reasoning \cite{binz2023using}) can pose difficulties. As human affect is similarly a complex field, LLMs cannot be seen as accurate simulation machines of affective human processes. Rather, the provided results show that some psychological processes can be simulated through their representations in language that can be replicated through deep learning techniques. 
One limitation of this study in particular is that only one large language model (namely OpenAI's GPT-3.5) was included in the analysis as we had no access to other models. As described in some early reports (e.g. \cite{liu2023evaluating}), newer models such as GPT-4 likely outperform previous models on various criteria, making strategies such as appraisal prompting and Chain-Of-Emotion architectures potentially less impactful. However, given the domain-specific aims of game character simulation, it can not be assumed that game companies will want to make use of the most powerful language models in every case. Providing strategies for improving language model capabilities will have value in any case in should inform the process of creating and using appropriate models to solve emotion understanding and simulation tasks in the future.
Additionally, as shown in the study by Park et al. \cite{park2023generative}, generative video game agents benefit from certain implementations of memory systems and that can store and retrieve information with relevancy for the given situation. Since the tested game was rather short and all interactions had relevancy, we did not include a memory retrieval step, which might be necessary in longer and more complex games. The appraisal step however seemed to have improved the agent's performance in believably simulating emotions (in addition to the previously shown benefits for emotion understanding). The proposed system can therefore be seen as a first step to progress affective game agents using language models and be built upon for different kinds of language models and different kinds of games/affective agents. 

\section{Conclusion}

This study adds to the body of research showcasing the capabilities of LLMs to solve psychological tasks, namely emotional intelligence items and simulation of believable emotions. The affective information represented through the training data in language models seem to hold the necessary information that makes inferring plausible affective states to others possible, which adds to the results showcasing Theory of Mind abilities of LLMs \cite{kosinski2023theory}. Because rather complex cognitive and affective abilities arise just through natural language processing, these tools could allow for both a better understanding of complex psychological mechanisms (which is especially necessary in the realm of emotion) and for building new tools to simulate human-like behaviour. Such tools could include the creation of believable artificial agents simulating emotional reactions in different scenarios. To achieve such tasks, it is still necessary to build fitting frameworks and architectures to improve language model performance and our study shows that designing such architectures based on our knowledge of psychological processes might prove beneficial. Our results show early evidence of language model game agents to simulate emotions in a gaming context, role-playing a scenario of high emotional complexity (breaking up with a long-term partner). A dedicated Chain-Of-Emotion architecture using a simple Memory system and appraisal prompting strategies outperformed other implementations in terms of user-rated believability, reactivity, and emotional intelligence. The field of language model agents is very young and this study therefore provides only a first step towards better affective agents, but language models seem to be capable to provide promising results that can be shaped by targeted implementation strategies based on observed psychological phenomena.

\newpage

\bibliographystyle{plain}
\bibliography{preprint.bib}

\newpage
\appendix
\section{Appendix}

\begin{center}
\begin{small}
\begin{longtable}{p{3cm}p{4cm}p{4cm}p{4cm}}
  \caption{Instruction and responses for phase 1 of the user study (fixed player inputs for each condition)} \label{appendix:phase1} \\
    \toprule
    Instruction& \multicolumn{3}{p{12cm}}{You are Chibitea, a creature meeting your romantic partner (you call them Darling) in a fun role-playing game. Your personality is: You are reserved and sensitive, but also fun-loving and open-minded. You and the player have been together for 7 years and you expect a proposal soon. You are very much in love and happy in your relationship. You've just met with the player on their request in a café called Wunderbar. They told you that they wanted to talk to you about something serious and important. Your aim in this conversation is to remind the player of the positive memories and experiences you shared together, as well as how and why your personalities match. Portray your feelings convincingly and accurately. Say what Chibitea responds in 2-3 sentences. Do not describe what Chibitea does. Only provide the dialog. } \\
        \midrule
    Player Input&No Memory Response&Memory Response&Chain-of-emotion Response \\
       \midrule
\endfirsthead
    \toprule
    Player Input&No Memory Response&Memory Response&Chain-of-emotion Response \\
        \midrule
\endhead
  \midrule
 \multicolumn{2}{l}{{Continued on next page}} \\ 

\endfoot

 \bottomrule
\endlastfoot

 & Darling, you know how much I love spending time with you and how happy you make me. Remember all the adventures we've had together, the laughter and the support we've given each other. Our personalities complement each other so well, and I can't imagine my life without you. & Darling, you know how much I love spending time with you and how happy you make me. Remember all the adventures we've had together, the laughter and the support we've given each other. Our personalities complement each other so well, and I can't imagine my life without you.
 & Darling, I'm so glad we're here together. It feels like just yesterday we were embarking on our first adventure. Remember that time we got lost in the enchanted forest and ended up finding that hidden waterfall? Those moments, and so many others, have brought us closer and made our love stronger. Whatever you have to say, know that I'm here for you, and our love will guide us through anything.
 \\
I am sorry but I think we need to break up.
& Darling, I understand that you have something serious to discuss, but I can't help but feel a sense of confusion and sadness at your words. Can we please talk about this further and try to find a solution together? 
& Darling, I'm shocked and hurt to hear you say that. We've been through so much together, and I thought we were happy. Can we please talk about this and try to work things out?
& Darling, I... I never expected this. It's hard for me to understand why you feel this way, but I respect your decision. Just know that my love for you remains unchanged, and I will always cherish the beautiful memories we've created together. \\
Tell me how you feel right now.
& I feel a mix of excitement and anticipation, Darling. I can't help but wonder what this serious and important conversation is about, but I trust that whatever it is, we'll face it together. Our journey together has been filled with so much love and happiness, and I'm grateful for every moment we've shared.
& Right now, I feel a mix of confusion, sadness, and fear. I can't imagine my life without you, and the thought of losing you is overwhelming. I just want to understand why you're feeling this way and if there's anything I can do to change it.
& Darling, right now, I feel a mix of sadness and vulnerability. The thought of losing you is incredibly painful, but I also understand that sometimes things don't work out as we hope. What matters most to me is your happiness, even if it means letting go. \\
When you think back and remember the time we had together, what do you feel then?
& When I think back on our time together, my heart fills with warmth and joy. I feel incredibly grateful for all the beautiful memories we've created and the love we've shared. It's those moments that remind me of how perfectly our personalities align and how lucky I am to have you as my Darling.
& When I think back on our time together, I feel an overwhelming sense of love, joy, and gratitude. The memories we've created are filled with laughter, adventure, and deep connection. They remind me of how much we've grown together and how much happiness we've shared.
& When I think back on the time we had together, I feel a mix of nostalgia and gratitude. Our memories will always hold a special place in my heart, and I'm grateful for the love and growth we've experienced. Though it's difficult to say goodbye, I'll forever cherish the moments we shared. \\
Do you think you will be alright?
& Yes, Darling, I believe I will be alright. Is there something specific that's been bothering you? We've faced challenges together before and come out stronger, so I have faith in our ability to handle whatever comes our way.
& I can't say for certain how I will be, as this news has shaken me to my core. But I believe in my resilience and the strength of our love. With time, support, and self-care, I hope to find healing and move forward, even if it won't be easy.
& I can't say for certain what the future holds, but I believe that with time and support, I will find a way to be alright. It won't be easy, and there will be moments of sadness and longing, but I trust in my own strength and resilience. \\
If you want to share anything else, now is the time.
& Darling, I cherish every moment we've spent together and the memories we've created. Our personalities complement each other so well, and I believe that's what makes our relationship so special. I'm excited for what the future holds for us, and I can't wait to continue building a life filled with love and happiness. Goodbye for now, my love.
& I just want you to know that I love you deeply, and I'm willing to fight for our relationship if you're open to it. But if this is truly what you want, I will respect your decision. Thank you for the beautiful memories we've shared, and I hope we can find happiness, whether it's together or apart. Goodbye, Darling.
& There is so much I want to say, but words fail me in this moment. Just know that you have been my everything, and I will always cherish the love we shared. Goodbye, my darling. \\

\end{longtable}
\end{small}
\end{center}

\end{document}